# ISLES'24 - A Real-World Longitudinal Multimodal Stroke Dataset

## Authors

Evamaria Olga Riedel[1], Ezequiel de la Rosa[2], The Anh Baran[1], Moritz Hernandez Petzsche[1], Hakim Baazaoui[3], Kaiyuan Yang[2], Fabio Antonio Musio[2,4], Houjing Huang[2], David Robben[5], Joaquin Oscar Seia[6], Roland Wiest[7,8], Mauricio Reyes[9,10], Ruisheng Su[11], Claus Zimmer[1], Tobias Boeckh-Behrens[1], Maria Berndt[1], Bjoern Menze[2], Daniel Rueckert[12,13], Benedikt Wiestler[1,14,15], Susanne Wegener[3,16*], Jan Stefan Kirschke[1*]

### Affiliations

[1] Department of Diagnostic and Interventional Neuroradiology, School of Medicine and Health, TUM Klinikum Rechts der Isar, Technical Universtiy of Munich.

[2] Department of Quantitative Biomedicine, University of Zurich, Zurich, Switzerland.

[3] Department of Neurology, University Hospital of Zurich and University of Zurich, Zurich, Switzerland.

[4] Center for Computational Health, Zurich University of Applied Sciences, Zurich, Switzerland

[5] icometrix, Leuven, Belgium.

[6] University of Girona, Girona, Spain.

[7] Support Center of Advanced Neuroimaging (SCAN), University Institute of Diagnostic and Interventional Neuroradiology, Inselspital, Bern, Switzerland.

[8] University Institute of Diagnostic and Interventional Neuroradiology, University Hospital Bern, Inselspital, University of Bern, Bern, Switzerland.

[9] ARTORG Center for Biomedical Research, University of Bern, Bern, Switzerland.

[10] Department of Radiation Oncology, University Hospital Bern, University of Bern, Switzerland.

[11] Department of Biomedical Engineering, Eindhoven University of Technology, Eindhoven, The Netherlands.

[12] Chair for AI in Healthcare and Medicine, Technical University of Munich (TUM) and TUM University Hospital, Munich, Germany

[13] Department of Computing, Imperial College London, London, United Kingdom

[14] TranslaTUM - Central Institute for Translational Cancer Research, Technical University of Munich, Munich, Germany.

[15] AI for Image-Guided Diagnosis and Therapy, School of Medicine and Health, Technical University of Munich, Munich, Germany

[16] University of Zurich, Zurich, Switzerland.

\* These authors contributed equally.

corresponding author: Evamaria O. Riedel (evamaria.riedel@tum.de)

# Abstract

Stroke remains a leading cause of global morbidity and mortality, imposing a heavy socioeconomic burden. Advances in endovascular reperfusion therapy and CT and MR imaging for treatment guidance have significantly improved patient outcomes. Developing machine learning algorithms that can create accurate models of brain function from stroke images for tasks like lesion identification and tissue survival prediction requires large, diverse, and well-annotated public datasets. While several high-quality image datasets in stroke exist, they include only single time point data. Data over different time points are essential to accurately identify lesions and predict prognosis. Here, we provide comprehensive longitudinal stroke data, including (sub-)acute CT imaging with angiography and perfusion, follow-up MRI after 2-9 days, and acute and longitudinal clinical data up to a three-month outcome. The dataset also includes vessel occlusion masks from acute CT angiography and delineated infarction masks in follow-up MRI. This multicenter dataset consists of 245 cases and is a solid basis for developing powerful machine-learning algorithms to facilitate clinical decision-making.

## Background & Summary

### Estimation of infarct growth from (sub-)acute imaging and influencing factors is crucial for clinical decision-making

Stroke is a leading cause of morbidity and mortality worldwide, imposing a substantial socioeconomic burden [1-5]. Over the past decade, the advent of endovascular reperfusion therapy has significantly improved outcomes for patients with significant vessel occlusions [6-8]. Utilization of computed tomography (CT) and magnetic resonance imaging (MRI) image-based guidance for reperfusion treatment decisions has even further advanced patient outcomes. This approach was integrated into clinical practices and endorsed by national guidelines. Clinical decisions regarding the treatment of ischemic stroke patients hinge on accurately estimating core (irreversibly damaged tissue) and penumbra (salvageable tissue) volumes [9]. The current clinical standard for estimating perfusion volumes involves deconvolution analysis, which entails generating perfusion maps through perfusion CT and MR deconvolution and applying thresholds to these maps [10]. However, variations in deconvolution algorithms, their technical implementations, and the thresholds used in software packages can significantly affect the estimated lesion sizes [11].

Furthermore, due to persistent hypoxia and ongoing, irreversible damage to penumbral tissue, the ischemic core tissue tends to expand over time. This expansion is unique to each patient and influenced by factors such as thrombus location and collateral circulation. Understanding the rate of core expansion and its determinants is crucial in clinical practice for evaluating the necessity of transferring a patient to a comprehensive stroke center and choosing treatment indications based on transportation times. Predicting infarct growth from underlying and influencing parameters can provide interventional radiologists with insights into the potential benefits of additional reperfusion attempts and is essential for informed clinical decision-making [12].

### Powerful datasets are essential to develop effective machine-learning algorithms to aid in clinical decision making

As machine-learning techniques continue to advance, powerful AI algorithms are emerging [13], and their integration into clinical practice is anticipated shortly. Automated detection and prediction of stroke lesions have the potential to standardize segmentation and provide automated support for therapeutic decisions, outcome forecasting, and stroke etiology classification in clinical settings. To create effective algorithms, well-annotated datasets are essential. Whereas previously, only a few datasets with (sub-)acute stroke data have been available, recently, powerful large datasets have been released. The previous version of the ISLES challenge dataset, e.g., includes FLAIR, DWI, and ADC from 400 patients [14]. A large stroke dataset (n= 1271), including single-channel T1-w images and delineated lesion masks, is proposed by Liew et al. [15]. A recent dataset from the Johns Hopkins Comprehensive Stroke Center by Liu et al. (n = 2888) includes MRI data with diffusion-weighted, fluid-attenuated, T1- and T2- weighted, perfusion-weighted, and susceptibility-weighted sequences [16]. Another recent dataset from a cohort from South Carolina by Absher et al. (n = 1715) offers MRI data with diffusion-weighted, fluid-attenuated, and T1-weighted sequences [17]. A first effort to build a paired dataset with non-contrast CT (NCCT) and apparent diffusion coefficient (ADC) maps was made by Gómez et al. [18] but consists of only a relatively small set (n = 36) and data only from a single time point.

**The first longitudinal and multimodal stroke dataset offers until now unprecedented possibilities to determine infarct lesion outcomes**
In contrast, our here-described dataset offers, as a first of its kind, the possibility to further investigate the rate of infarct core expansion over time by providing longitudinal imaging at two time points: (sub-)acute CT imaging before therapy and follow-up MRI 2-9 days after intracranial reperfusion therapy. NCCT may reveal infarct areas that are not visible in CT perfusion (CTP, e.g., in patients with spontaneous reperfusion). CT angiography (CTA) can offer information on thrombus location, influencing infarct growth. Complemented is longitudinal clinical data up to a three-month outcome, which enables taking into account patient-specific characteristics like demographics, pre-existing risk factors, medication, and the type of referral or intervention times. Additionally, the diverse composition of the dataset, combined with clinical and demographic tabular data, also allows integration with other datasets, like, e.g., the acute and early sub-acute ischemic stroke cohort from the Johns Hopkins Comprehensive Stroke Center [16].

Data preparation, which is briefly illustrated in **Figure 1** and described in detail in the methods section, led to a dataset of 245 cases divided into a publicly available train set (n = 149) and - to ensure fair and continuous evaluation of machine learning algorithms - a hidden test set (n = 96). For each case, the following is provided:
- (Sub-)acute (baseline) CT imaging including NCCT, CTA provided with a segmentation of the point of vessel occlusion and a segmentation of the Circle of Willis as well as CTP with derived perfusion maps of CBF (Cerebral Blood Flow), CBV (Cerebral Blood Volume), MTT (Mean Transit Time) and Tmax (Time-to-maximum of the residue function).
- Outcome MR imaging comprises diffusion-weighted imaging (DWI), ADC maps, and DWI lesion segmentations.
- Baseline and outcome clinical parameters as well as intervention times.

Our dataset aims to facilitate algorithm development for clinical and research purposes, particularly for lesion identification and prognosis by predicting post-interventional stroke outcomes, aiding in clinical decision-making. Computational challenges like VerSe [19], BRATS [20], or ISLES [14,21,22] have proven to be an effective trigger for accelerating such developments. The here-presented dataset was the basis for the 2024 edition of the Ischemic Stroke Lesion Segmentation (ISLES) Challenge (https://www.isles-challenge.org/), which continuously aims to establish benchmark methods for acute and sub-acute ischemic stroke lesion segmentation, aiding in creating open stroke imaging datasets and evaluating cutting-edge image processing algorithms. ISLES'24 aims to benchmark algorithms for final, post-treatment infarct segmentation, solely utilizing multimodal pre-treatment CT data treatments [23]. Such outputs may enrich or guide clinical decision-making in reperfusion treatments.

## Methods

### Ethical statement

This multi-center study used data from studies approved by their local ethics committees. It was executed in agreement with the ethical standards of the 1964 Declaration of Helsinki and its updated version [24]. Due to defacing and rigorous anonymization, the ethics committee at the receiving site (University of Zurich) approved sharing the de-identified data.

### Patient selection

Patients 18 years or older who underwent a (sub-)acute CT stroke imaging protocol followed by intracranial interventional reperfusion therapy and follow-up MR imaging of the brain for suspected acute or sub-acute stroke were included in this study. The (sub-)acute CT protocol consisted of NCCT, CTA, and CTP. MR imaging occurred 2-9 days after CT imaging and subsequent intracranial interventional reperfusion therapy. At the minimum, it consisted of a Fluid attenuated inversion recovery (FLAIR) sequence, DWI consisting of a trace image at a b-value up to 1000 s/mm$^2$, and a corresponding ADC map. Acquired images are shown exemplarily in **Figure 2**. To minimize random effects of treatment success, exclusively patients with an almost complete recanalization, rated as modified treatment in cerebral infarction (mTICI) 2c (near complete perfusion except for slow flow or distal emboli in a few distal cortical vessels [25]) or 3 (complete antegrade reperfusion of the previously occluded target artery ischemic territory, with absence of visualized occlusion in all distal branches [26]), were included in this study.

### Image acquisition

Healthcare professionals obtained images as part of the clinical imaging routine for stroke patients at two stroke centers in Germany and Switzerland: Center 1 - University Hospital of the Technical University of Munich in Munich, Germany, and Center 2 - University Hospital of Zurich in Zurich, Switzerland. CT image acquisition was performed on the following devices: Somatom Force, Somatom Xcite (Siemens Healthcare), Somatom AS+ (Siemens), Brilliance 64, and Ingenuity (Philips Healthcare). MR Image acquisition was carried out on 3 T Philips MRI scanners (Achieva, Ingenia), a 3 T Siemens MAGNETOM MRI scanner (Verio, Trio), or 1.5 T Siemens MAGNETOM MRI scanners (Avanto). Stemming from two centers and different scanner models and manufacturers, the dataset described here allows the development of robust and generalizable stroke lesion segmentation algorithms. Additionally, the scans were intentionally chosen to be heterogeneous in lesion size, quantity, and location to guarantee the best possible and generalized training of the algorithms.

### Data pre-processing

All medical imaging files were exported from the Picture Archiving and Communication System (PACS) in the NIfTI format. The imaging data underwent irreversible anonymization before releasing the dataset and complied with the ethical approval acquired for this challenge. All images were released as NIfTI (Neuroimaging Informatics Technology Initiative, https://nifti.nimh.nih.gov/nifti-1) files using the BIDS (Brain Imaging Data Structure, https://bids.neuroimaging.io) convention [27].

CT scans were defaced using in-house developed scripts based on TotalSegmentator [28]. Data pre-processing consisted of image co-registration to compensate for head motion and temporal resampling (1 frame/second) of the 4D CTP series. Then, perfusion maps (CBF, CBV, MTT, and Tmax) were derived from the 4D CTP series using the clinical, FDA-cleared software icobrain cva [29,30]. CTA, CTP (including derived perfusion maps), and DWI/ADC scans were linearly co-registered to the NCCT space using rigid transformations for CT-based images, and affine transformations for MRI. Registrations were performed using Elastix [31] and NiftyReg [32]. MRIs were skull-stripped using HD-BET [33]. All images are released 'raw' (i.e., solely anonymized and defaced) and preprocessed (i.e., resampled and co-registered to the NCCT space).

**Vessel occlusion segmentation in CTA scans**
Various factors influence infarct core expansion, including individual patient attributes, collateral circulation, and thrombus location. When determining the sensibility of an intervention, it's crucial to consider not only which tissue can be saved and how the patient might benefit but also the location of the occlusion. This location significantly impacts whether mechanical therapy is feasible and low-risk. Consequently, our dataset includes masks for the CTA scans that indicate the point of vessel occlusion—the final segment of the vessel before the occlusion is marked. If there are tandem cases (e.g., in the internal carotid artery and middle cerebral artery), both are marked. In cases of multiple occlusions (e.g., in the anterior and middle cerebral territories), these are also marked. If a large thrombus creates a filling defect in the contrast medium proximally, followed by a vessel that is briefly filled with contrast, both the proximal end of the thrombus and the distal end of the obstruction are delineated. Since the CTAs were cropped uniformly and cover only the head and not the neck, occlusions of the internal carotid artery (especially near the carotid bifurcation) may not be captured. In such instances, the point where the internal carotid artery would be, is marked on the lowest available slice. Segmentations were carried out manually by a neuroradiology resident using ITK-SNAP [34] (www.itksnap.org), controlled and revised if necessary by an experienced neuroradiology attending. An example of the segmentations is shown in **Figure 3A**.

**Ground Truth stroke lesion segmentation**
A hybrid human-algorithm annotation scheme to segment all cases was used. MRI input data was anonymized by conversion to NIfTI format, which agreed with the BIDS. We used the deep-learning ensemble model from ISLES'22 [35] to segment final infarcts in post-treatment DWI. An experienced neuroradiologist from the University Hospital of Munich (TUM Clinic) or the University Hospital of Zurich (UZH) visually inspected the resulting masks. Cases with annotations of suboptimal quality were manually revised by a medical student with special stroke lesion segmentation training, followed by a neuroradiology resident using ITK-SNAP software [34] (www.itksnap.org). An example is shown in **Figure 3B**.

**Segmentation of the Circle of Willis (CoW) for CTA scans**

The CoW segmentation baseline is a two-stage approach. In the first stage, the CoW region is localized in the input CTA, and a bounding box is predicted. The bounding box is then slightly enlarged and cropped as input to the second stage. Stage two is a segmentation model for the cropped image. Both models have been trained on a combined CTA and MRA dataset using an extended U-net [36] framework. Training data were augmented by cross-registering the CT and MR modalities of the same patients. Deep supervision and mirroring in data augmentation were turned off during training. A balanced cross entropy (CE) loss weight and a higher bias to sample the foreground were employed. The loss was a combined CE, Dice, Skeleton Recall [37], and a customized topology loss. The inference was an ensemble of three models. These segmentations serve as a *silver* ground-truth, offering a coarse reference segmentation not traced by experts. An example is shown in **Figure 4**. Further details can be found in the TopCoW (Topology-Aware Anatomical Segmentation of the Circle of Willis) summary [38].

**Demographics**

Given that infarct core expansion and the decision for mechanical therapy are influenced by various factors, including individual patient attributes, clinical tabular data is provided alongside the images for each subject. This tabular data includes age and sex distribution, medical history (atrial fibrillation, hypertension, diabetes mellitus, hyperlipidemia), medication (anticoagulation, statins, platelet aggregation inhibitors), laboratory values (glucose, leucocytes, CRP, INR), type of referral (wake-up, in-house, referral from external clinic), times (onset to door, alert to door, door to imaging, door to groin, door to first series, time of intervention, door to recanalization), clinical scores like National Institutes of Health Stroke Scale (NIHSS) (at admission, after 24 hours, at discharge) and modified Rankin Scale (mRS) (at admission, premorbid, at discharge, at three months) and the outcome of recanalization (mTICI postinterventional). Means and standard deviations are provided in **Table 1**. Parameters are divided into outcome parameters after recanalization (mTICI postinterventional, NIHSS and mRS at 24 hours, NIHSS and mRS at discharge, mRS at three months) and baseline parameters before intracranial interventional recanalization (all others). Laboratory values and times were randomly altered by ± 5 % for anonymization purposes.

Further information regarding the clinical data is provided in the info table provided in the data repository.

| Parameter | Train | Test | All (Train + Test) |
|---|---|---|---|
| **Patients** | 149 | 96 | 245 |
| Center 1 | 67% | 52% | 61% |
| Center 2 | 33% | 48% | 39% |
| **Age** | 71.9 ± 14.4 | 69.9 ± 15.2 | 71.1 ± 14.7 |
| **Gender** | | | |
| Female | 52% | 44% | 49% |
| Male | 44% | 56% | 51% |
| **Medical history** | | | |
| Atrial fibrillation | 29% | 34% | 31% |
| Hypertension | 59% | 65% | 61% |
| Diabetes mellitus | 17% | 16% | 17% |
| Dyslipidemia | 30% | 45% | 36% |
| **Medication** | | | |
| Anticoagulation | 18% | 17% | 18% |
| Lipid lowering drugs | 27% | 37% | 31% |
| Platelet aggregation inhibitors | 25% | 28% | 26% |
| **Laboratory values** | | | |
| Glucose (mg/dl) | 109.7 ± 38.8 | 107.5 ± 53.5 | 108.8 ± 45.2 |
| Leucocytes (G/l) | 8.9 ± 3.1 | 9.6 ± 4.3 | 9.2 ± 3.5 |
| CRP (mg/dl) | 1.0 ± 1.7 | 1.8 ± 2.7 | 1.3 ± 2.1 |
| INR | 1.0 ± 0.2 | 1.1 ± 0.3 | 1.0 ± 0.3 |
| **Type of referral** | | | |
| Wake-up | 30% | 23% | 27% |
| In-house | 5% | 2% | 4% |
| Referral from external clinic | 12% | 12% | 12% |
| **Baseline clinical parameters** | | | |
| mRS premorbid | 0.9 ± 1.3 | 0.9 ± 2.1 | 0.9 ± 1.6 |
| mRS at admission | 4.0 ± 1.1 | 4.1 ± 1.1 | 4.0 ± 1.1 |
| NIHSS at admission | 11.5 ± 6.2 | 12.9 ± 6.3 | 12.0 ± 6.3 |
| **Times** | | | |
| Onset to door | 03:40 ± 04:53 | 03:03 ± 03:32 | 03:28 ± 04:29 |
| Alert to door | 00:34 ± 00:26 | 00:43 ± 00:22 | 00:37 ± 00:24 |
| Door to imaging | 00:25 ± 00:46 | 00:24 ± 00:41 | 00:25 ± 00:44 |
| Door to groin | 01:18 ± 00:48 | 01:45 ± 02:39 | 01:28 ± 01:45 |
| Door to first series | 01:27 ± 00:45 | 01:31 ± 0:54 | 01:28 ± 00:48 |
| Time of intervention | 00:48 ± 00:44 | 00:39 ± 00:26 | 00:45 ± 00:39 |
| Door to recanalization | 02:17 ± 01:07 | 02:08 ± 01:11 | 02:14 ± 01:08 |
| **Outcome clinical parameters** | | | |
| mTICI postinterventional 2c | 20% | 10% | 17% |
| mTICI postinterventional 3 | 80% | 90% | 83% |
| NIHSS after 24 hours | 6.0 ± 5.7 | 6.7 ± 5.6 | 6.2 ± 5.7 |
| NIHSS at discharge | 5.2 ± 7.2 | 4.3 ± 6.3 | 4.9 ± 6.9 |
| mRS at 24 h | 3.4 ± 1.5 | 3.4 ± 1.7 | 3.4 ± 1.6 |
| mRS at discharge | 2.4 ± 1.8 | 2.5 ± 1.7 | 2.4 ± 1.8 |
| mRS at three months | 2.1 ± 2.1 | 1.8 ± 2.1 | 2.0 ± 2.1 |

**Table 1:** Demographics from our test and training set. Values are either given in percent or as mean ± standard deviation.

## Data Records

### Data repository and storage

The complete training data set (n = 149) has been publicly made available under the Creative Commons license CC BY-SA-NC 4.0 (Attribution-NonCommercial) and can be accessed via the Challenge website (https://isles-24.grand-challenge.org/dataset/). The test set is kept hidden to guarantee fair and ongoing evaluation of machine learning algorithms.

### Data structure

The data is separated into a training dataset (n = 149 cases) and a test dataset (n = 96 cases). The train set contains n = 99 scans from Center 1 and n = 50 cases from Center 2 and is publicly available. The test set contains n = 50 scans from Center 1 and n = 46 scans from Center 2 and serves as a hidden test set in the ISLES'24 challenge [23]. The training and test datasets have been divided to ensure that the training and the data sets include a similar range of stroke lesion patterns, from extensive territorial infarcts to minor punctate ischemia. A summary of infarct volume per scan (scan-level infarct volume), infarct volume per unconnected lesion (lesion-level infarct volume), and the number of unconnected infarcts per scan and center are provided in **Table 2**. Unconnected infarcts are calculated through connected components analysis using the Python library cc3d [39].

| Center | Parameter | Scan infarct volume (ml) | Number of unconnected infarcts | Lesion-wise infarct volume (ml) |
|---|---|---|---|---|
| 1 | Mean (std) | 32.88 (50.39) | 12.23 (12.50) | 2.62 (16.27) |
|  | [min, max] | [0.12, 318.25] | [1.00, 84.00] | [0.00, 318.15] |
| 2 | Mean (std) | 33.76 (46.46) | 7.96 (8.09) | 4.10 (19.08) |
|  | [min, max] | [1.33e-3, 259.63] | [1.00, 53.00] | [0.00, 259.63] |
| All | Mean (std) | 33.16 (48.67) | 10.55 (11.10) | 3.05 (17.11) |
|  | [min, max] | [1.33e-3, 318.25] | [1.00, 84.00] | [0.00, 318.15] |

**Table 2:** Summary of infarct lesion statistics. Std = standard deviation, min = minimum, max = maximum.

Infarct statistics by type of stroke pattern are shown in **Table 3A**. The scans are categorized into one of four clinical sub-groups based on the type of lesion and stroke pattern, as defined in de la Rosa et al. [24]. These groups included scans with no final infarcts, scans showing a single vessel infarct, where the largest lesion accounted for more than 95% of the total lesion volume; scans with scattered infarcts due to micro-occlusions, defined by three or more individual lesions, with the largest lesion comprising less than 60% of the total lesion volume, or the total volume being under 5 ml; and finally, scans with a single vessel infarct accompanied by scattered infarcts, which included all remaining cases.

The dataset displays lesions in the vascular territories supplied by predominantly the anterior and middle cerebral arteries and a good 10 % posterior circulation, exhibiting significant spatial and anatomical variability in DWI final infarct lesions. **Table 3B** presents infarction statistics for each affected brain anatomical structure. The MRI scans were first segmented into eleven anatomical structures using a deep-learning model [28] for this analysis. Subsequently, the scans were assigned to a specific anatomical region based on the structure with the largest infarct volume.

| | | Center 1 | Center 2 | All |
|---|---|---|---|---|
| **A) Stroke pattern** | **No final infarct** | 3 (2.0 %) | 0 (0.0 %) | 3 (1.2 %) |
| | **Scattered infarcts based on micro-occlusions** | 52 (35.4 %) | 28 (29.8 %) | 81 (33.1 %) |
| | **Single vessel infarct** | 55 (37.4 %) | 42 (44.7 %) | 98 (40.0 %) |
| | **Single vessel infarct with accompanying scattered infarcts** | 37 (25.2 %) | 24 (25.5 %) | 63 (25.7 %) |
| **B) Anatomical location** | **Brainstem** | 17 (11.5 %) | 3 (3.2 %) | 20 (8.2 %) |
| | **Caudate nucleus** | 3 (2.0 %) | 1 (1.1 %) | 4 (1.6 %) |
| | **Cerebellum** | 4 (2.7 %) | 3 (3.20 %) | 7 (2.9 %) |
| | **Frontal lobe** | 40 (27.0 %) | 28 (29.8 %) | 68 (27.8 %) |
| | **Insular cortex** | 1 (0.7 %) | 0 (0.0 %) | 1 (0.4 %) |
| | **Internal capsule** | 1 (0.7 %) | 1 (1.1 %) | 2 (0.8 %) |
| | **Lentiform nucleus** | 4 (2.7 %) | 1 (1.1 %) | 6 (2.4 %) |
| | **Occipital lobe** | 3 (2.0 %) | 2 (2.1 %) | 5 (2.0 %) |
| | **Parietal lobe** | 57 (38.8 %) | 38 (40.4 %) | 97 (39.6 %) |
| | **Temporal lobe** | 17 (11.6 %) | 15 (16.0 %) | 33 (13.5 %) |
| | **Thalamus** | 0 (0.0 %) | 2 (2.1 %) | 2 (0.8 %) |

**Table 3:** Infarction statistics by type of stroke pattern (A) and anatomical location (B).

## Folder structure

The data is divided into test and train sets as described above and released in BIDS format. Each set is subdivided into three folders: raw data, derivatives, and phenotypes. Each of these three folders is divided into two sessions for every case: Session 1(ses-01), which refers to the (sub-)acute CT imaging, and Session 2 (ses-02), which relates to the follow-up MR imaging.

The raw data folder includes raw CT and MR images. In ses-01 includes the raw NCCT (Identifier_ses-01_ncct.nii.gz), the raw CTA image (Identifier_ses-01_cta.nii.gz), and the raw CTP image (Identifier_ses-01_ctp.nii.gz) as well as a folder with raw perfusion maps (CBV, CBF, MTT, and Tmax; e.g., Identifier_ses-01_tmax.nii.gz). Ses-02 provides raw DWI (Identifier_ses-02_dwi.nii.gz) and ADC map (Identifier_ses-02_adc.nii.gz).

In the derivatives folder, ses-01 includes the preprocessed CTA image, the preprocessed CTP image, and the preprocessed perfusion maps, following the same folder structure as for raw data. For all preprocessed data, the term

'space-ncct' is added in the naming (e.g., Identifier_ses-01_space-ncct_cta.nii.gz). Additionally, the CTA vessel occlusion segmentation mask (Identifier_ses-01_msk.nii.gz) is provided. Ses-02 includes the stroke infarct lesion mask for DWI and ADC (Identifier_ses-02_lesion-msk.nii.gz).

In the phenotype folder, clinical tabular data is provided. Ses-01 includes baseline clinical data (Identifier_ses-01_demographic_baseline.csv) and ses-02 outcome clinical data (Identifier_ses-02_outcome.csv). An info sheet, included with our data in the repository, defines the clinical baseline and outcome parameters. It also lists all cases with their respective originating centers and the sets they are assigned to.

## Technical Validation

The medical imaging data presented here was sourced from the PACS of the respective institutions, ensuring full compliance with the legal standards and quality controls for medical imaging acquisition in Germany, the European Union, and Switzerland. It also adheres to the industrial standards set by the scanner vendors. Our objective was to curate a dataset that reflects real-world stroke scenarios. Therefore, only cases with severe motion artifacts, making the images unfit for diagnostic use, were excluded from the dataset. We did not exclude cases based on other quality concerns, such as signal loss or spatial distortions, relying on the imaging standards maintained in clinical practice at the participating centers. Additionally, no preference was given to cases based on whether they were acquired at 1.5 T or 3 T.

### Inter-rater Analysis

The inter-rater agreement analysis of the infarct stroke segmentation against two expert raters was evaluated in Hernandez Petzsche et al. [14] using the Dice similarity coefficient and volume difference metrics. When comparing the segmentations to external rater I, a Dice score of 0.90 ± 0.09 and a volumetric difference of 2.37 ± 2.59 ml were obtained. The comparison with external rater II showed a Dice score of 0.86 ± 0.13 and a stroke infarct volume difference of 6.56 ± 13.37 ml [14].

## Code Availability

To help future users familiarize themselves with the images, we have published the ISLES 2024 GitHub repository at https://github.com/ezequieldlrosa/isles24. This repository includes scripts for image reading, visualization, and performance quantification using the metrics employed in the ISLES'24 challenge [23] to rank participants.

## Acknowledgments

None.

## Author contributions

EOR manuscript preparation, data preparation and review, segmentation preparation and review, clinical data acquisition and preparation, and image rating. EdlR code development, project design, data review, and manuscript preparation. TAB, MHP image rating, lesion segmentation. HB data acquisition, data review. KY, DR, JOS code development. MR, BM, RW, RS project design. SW data acquisition, project design. CZ, TBB, MB data acquisition, manuscript review. BW code development, project design, manuscript review. JSK code development, data acquisition, segmentation correction, project design, data review, and manuscript preparation. All authors revised and approved of the final manuscript.

## Funding

EOR is supported by the TUM KKF Clinician Scientist Program, project reference E-19 and H-27. EdlR and BM are supported by the Helmut Horten Foundation. HB is supported by the Koetser Foundation and the 'Young Talents in Clinical Research' program of the SAMS and of the G. & J. Bangerter-Rhyner Foundation. SW is supported by the Swiss National Science Foundation (Grant No. 310030_200703), the UZH Clinical Research Priority Program (CRPP) Stroke, and the Swiss Heart Foundation.

## Competing interests

Independent of this work, TBB consults for MicroVention, Balt, and Acandis and has received speaker honoraria from Philips and Phenox. CZ disclosed no relevant relationships regarding activities related to this article but has served on scientific advisory boards for Philips and Bayer Schering, is a co-editor on the Advisory Board of Clinical Neuroradiology, and has received speaker honoraria from Bayer-Schering and Philips. CZ's institution has received research support and investigator fees for clinical studies from numerous companies, including Biogen Idec, Quintiles, MSD Sharp & Dome, Boehringer Ingelheim, and others. SW has received speaker honoraria from Springer and Teva Pharma and consultancy fees from Bayer and Novartis. JOS is employed by Methinks AI. BW has received speaker honoraria from Philips. JSK is the cofounder of Bonescreen GmbH, which is unrelated to this work. JSK has received speaker honoraria from Novartis and is a shareholder of Bonescreen GmbH, which is unrelated to this work. EdlR was employed by icometrix. All other authors declare no conflicts of interest.

# Figures

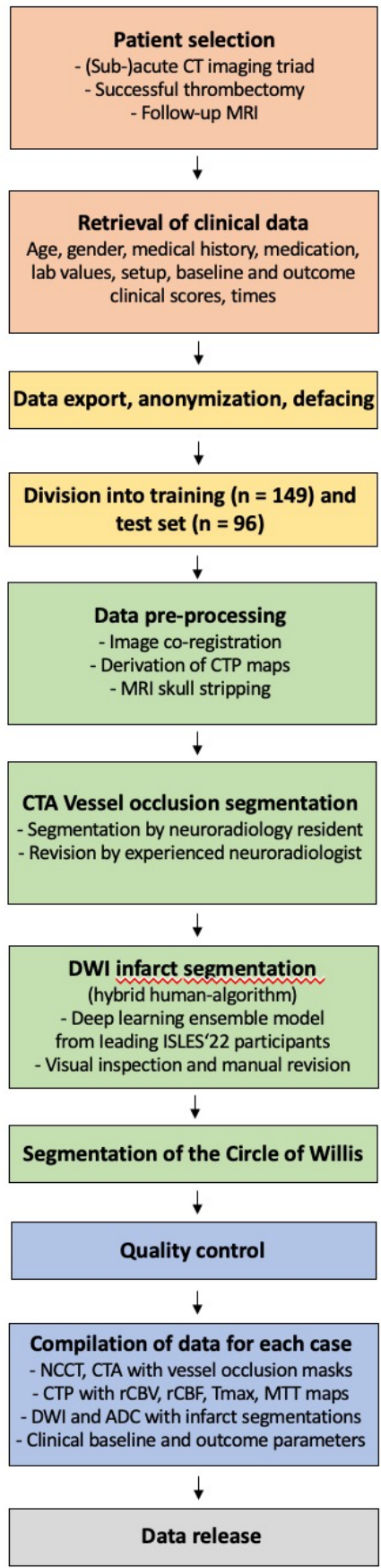

**Figure 1:** Workflow to generate the longitudinal multimodal (sub-)acute stroke dataset.

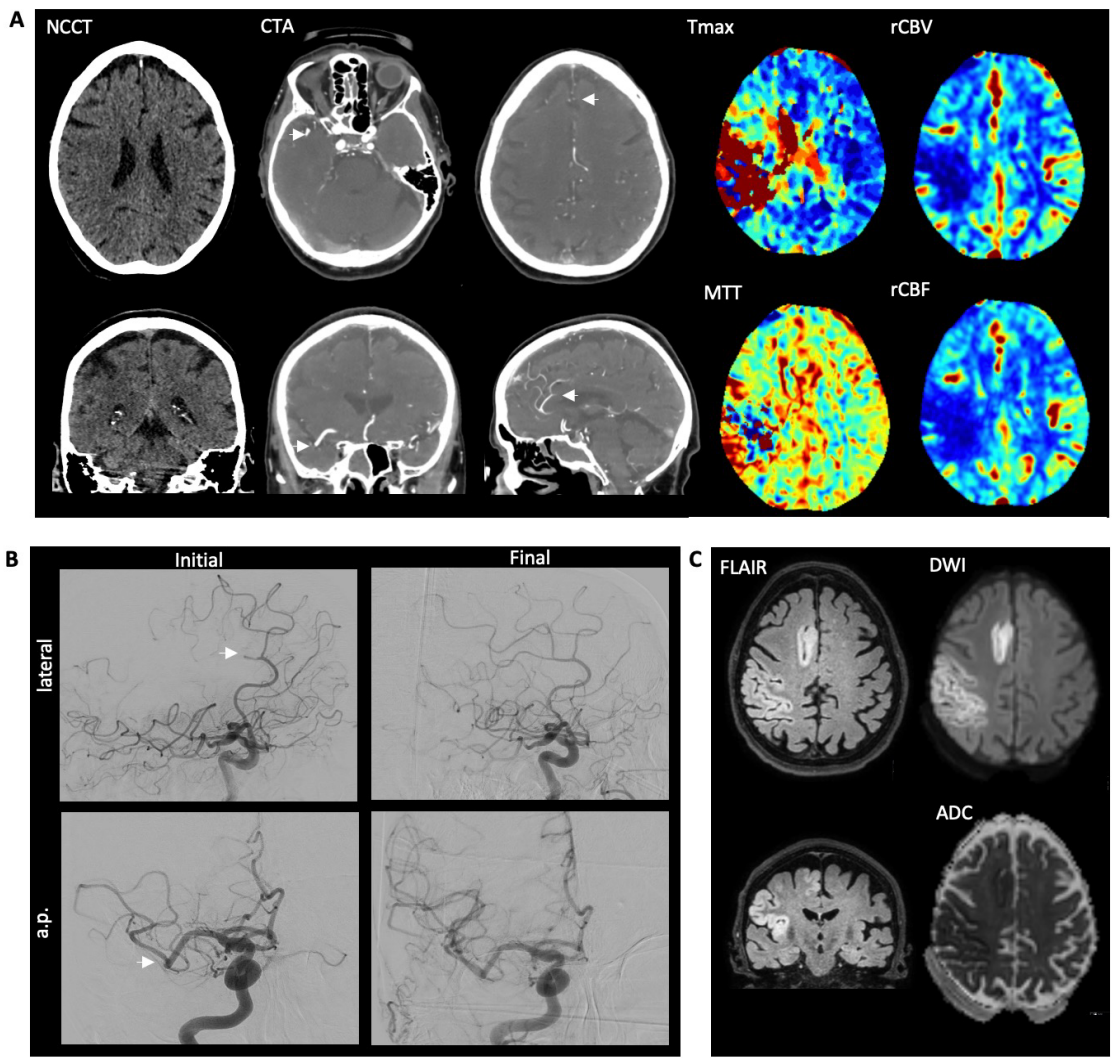

**Figure 2:** An example of images acquired for our longitudinal stroke dataset featuring a patient who underwent acute and follow-up stroke imaging. The patient experienced a sudden collapse followed by left-sided weakness. Upon arrival at our stroke unit, neurologists identified significant left-sided hemiparesis, rightward head and gaze deviation, dysarthria, left-sided neglect, and an NIHSS score of 17. **(A)** Initial CT imaging, performed approximately 1 hour and 20 minutes after symptom onset, revealed infarction in the right anterior cerebral artery (ACA) and the right middle cerebral artery (MCA) territories with matching vessel occlusions and perfusion deficits. The CT scan itself took about 10 minutes. **(B)** Right ACA and MCA occlusions in Digital subtraction angiography before the intracranial intervention, which commenced around 50 minutes after the initial CT scan and lasted approximately 1 hour and 30 minutes. **(C)** Follow-up MRI imaging was conducted 4 days after the initial acute CT imaging.

For all patients in the dataset, provided imaging includes NCCT, CTA, and CTP with derived maps (rCBV, rCBF, Tmax, MTT), along with DWI and ADC acquired 2-9 days after the initial imaging for each case, together with CTA segmentations of the point of vessel occlusion, segmentation of the circle of Willis and segmentation and lesion segmentation for DWI and ADC.

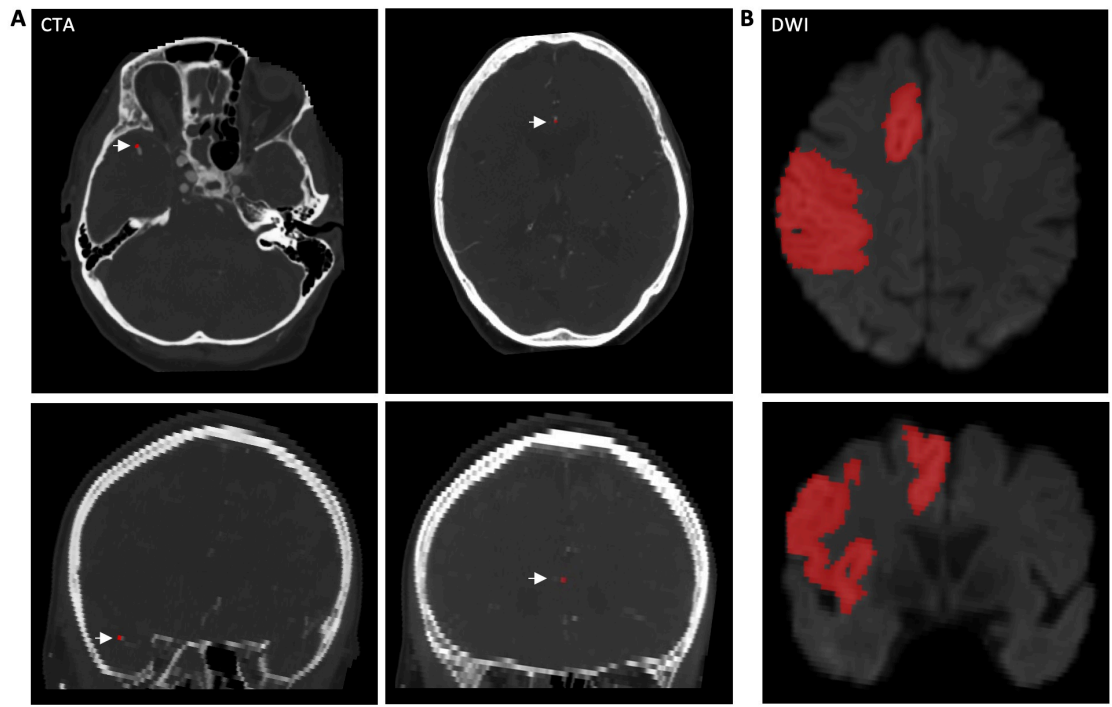

**Figure 3:** An example of the segmentations provided with each case, corresponding to the patient in Figure 2 with vessel occlusions in the right anterior cerebral artery and the right middle cerebral artery. On the left **(A),** an example of segmentation of the point of vessel occlusions (MCA and ACA on the right side) in the CTA is shown. On the right **(B),** an example of lesion segmentation for DWI is depicted (anterior and media cerebral artery territory on the right side).

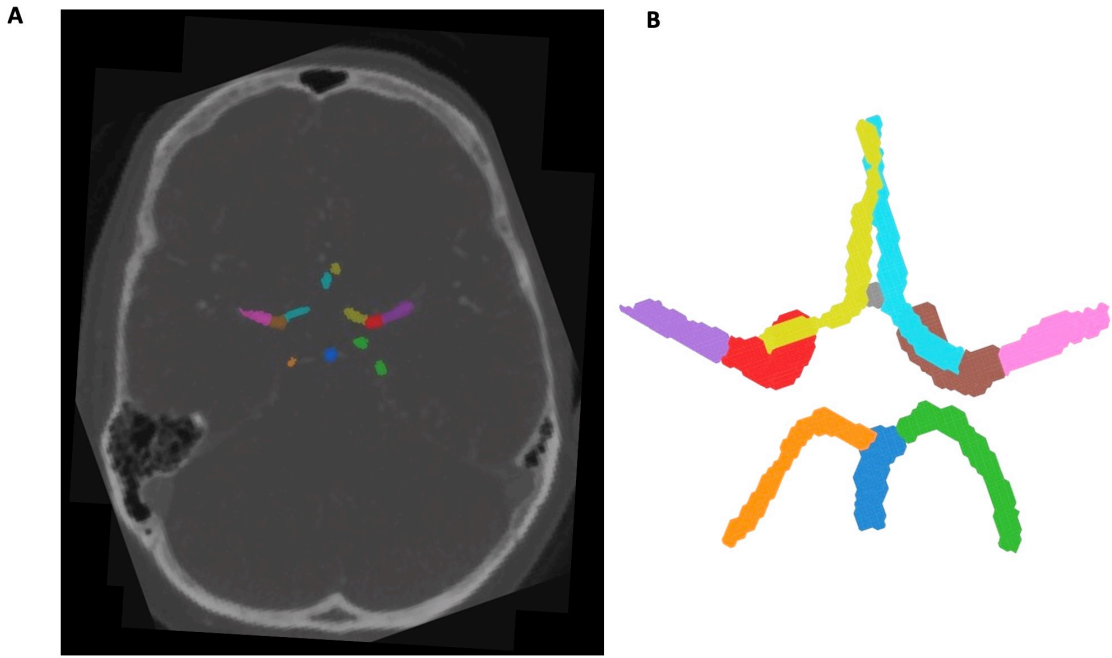

**Figure 4:** Segmentation of the Circle of Willis using TopCoW (Topology-Aware Anatomical Segmentation of the Circle of Willis). (**A**) Segmentation in an example axial slice. (**B**) 3D reconstruction of the Circle of Willis, view from below. Yellow = left ACA, light blue = right ACA, gray = AcomA, red = left ICA, purple = left MCA, brown = right ICA, pink = right MCA, orange = left PCA, green = right PCA, medium blue = basilar artery.

**Tables**

Tables and table legends are provided within the Word document.

# References

1 Lozano, R. *et al.* Global and regional mortality from 235 causes of death for 20 age groups in 1990 and 2010: a systematic analysis for the Global Burden of Disease Study 2010. *Lancet* **380**, 2095-2128, doi:10.1016/s0140-6736(12)61728-0 (2012).

2 GBD Stroke Collaborators. Global, regional, and national burden of stroke and its risk factors, 1990-2019: a systematic analysis for the Global Burden of Disease Study 2019. *Lancet Neurol* **20**, 795-820, doi:10.1016/s1474-4422(21)00252-0 (2021).

3 Centers for Disease Control and Prevention. Stroke Facts, < https://www.cdc.gov/stroke/data-research/facts-stats/index.html> (2024).

4 Centers for Disease Control and Prevention. National Center for Health Statistics Mortality Data on CDC WONDER <https://wonder.cdc.gov/mcd.html> (2024).

5 Benjamin, E. J. *et al.* Heart Disease and Stroke Statistics-2019 Update: A Report From the American Heart Association. *Circulation* **139**, e56-e528, doi:10.1161/cir.0000000000000659 (2019).

6 Berkhemer, O. A. *et al.* A randomized trial of intraarterial treatment for acute ischemic stroke. *N Engl J Med* **372**, 11-20, doi:10.1056/NEJMoa1411587 (2015).

7 Goyal, M. *et al.* Randomized assessment of rapid endovascular treatment of ischemic stroke. *N Engl J Med* **372**, 1019-1030, doi:10.1056/NEJMoa1414905 (2015).

8 Jovin, T. G. *et al.* Thrombectomy within 8 hours after symptom onset in ischemic stroke. *N Engl J Med* **372**, 2296-2306, doi:10.1056/NEJMoa1503780 (2015).

9 Albers, G. W. *et al.* Thrombectomy for Stroke at 6 to 16 Hours with Selection by Perfusion Imaging. *N Engl J Med* **378**, 708-718, doi:10.1056/NEJMoa1713973 (2018).

10 Lin, L., Bivard, A., Krishnamurthy, V., Levi, C. R. & Parsons, M. W. Whole-Brain CT Perfusion to Quantify Acute Ischemic Penumbra and Core. *Radiology* **279**, 876-887, doi:10.1148/radiol.2015150319 (2016).

11 Fahmi, F. *et al.* Differences in CT perfusion summary maps for patients with acute ischemic stroke generated by 2 software packages. *AJNR Am J Neuroradiol* **33**, 2074-2080, doi:10.3174/ajnr.A3110 (2012).

12 Robben, D. *et al.* Prediction of final infarct volume from native CT perfusion and treatment parameters using deep learning. *Med Image Anal* **59**, 101589, doi:10.1016/j.media.2019.101589 (2020).

13 Marcus, A. *et al.* Deep learning biomarker of chronometric and biological ischemic stroke lesion age from unenhanced CT. *NPJ Digit Med* **7**, 338, doi:10.1038/s41746-024-01325-z (2024).

14 Hernandez Petzsche, M. R. *et al.* ISLES 2022: A multi-center magnetic resonance imaging stroke lesion segmentation dataset. *Sci Data* **9**, 762, doi:10.1038/s41597-022-01875-5 (2022).

15 Liew, S. L. *et al.* A large, curated, open-source stroke neuroimaging dataset to improve lesion segmentation algorithms. *Sci Data* **9**, 320, doi:10.1038/s41597-022-01401-7 (2022).

16 Liu, C. F. *et al.* A large public dataset of annotated clinical MRIs and metadata of patients with acute stroke. *Sci Data* **10**, 548, doi:10.1038/s41597-023-02457-9 (2023).

17 Absher, J. *et al.* The stroke outcome optimization project: Acute ischemic strokes from a comprehensive stroke center. *Sci Data* **11**, 839, doi:10.1038/s41597-024-03667-5 (2024).

18 Gómez, S. *et al.* APIS: a paired CT-MRI dataset for ischemic stroke segmentation - methods and challenges. *Scientific reports* **14**, 20543, doi:10.1038/s41598-024-71273-x (2024).

19 Sekuboyina, A. *et al.* VerSe: A Vertebrae labelling and segmentation benchmark for multi-detector CT images. *Med Image Anal* **73**, 102166, doi:10.1016/j.media.2021.102166 (2021).

20    Bakas, S. *et al.* Identifying the Best Machine Learning Algorithms for Brain Tumor Segmentation, Progression Assessment, and Overall Survival Prediction in the BRATS Challenge. *arXiv:1811.02629v3* (2019).

21    Maier, O. *et al.* ISLES 2015 - A public evaluation benchmark for ischemic stroke lesion segmentation from multispectral MRI. *Med Image Anal* **35**, 250-269, doi:10.1016/j.media.2016.07.009 (2017).

22    Winzeck, S. *et al.* ISLES 2016 and 2017-Benchmarking Ischemic Stroke Lesion Outcome Prediction Based on Multispectral MRI. *Front Neurol* **9**, 679, doi:10.3389/fneur.2018.00679 (2018).

23    de la Rosa, E. et al. ISLES'24: Improving final infarct prediction in ischemic stroke using multimodal imaging and clinical data. *arXiv:2408.10966* (2024).

24    World Medical Association Declaration of Helsinki: ethical principles for medical research involving human subjects. *Jama* **310**, 2191-2194, doi:10.1001/jama.2013.281053 (2013).

25    Zaidat, O. O. *et al.* Recommendations on angiographic revascularization grading standards for acute ischemic stroke: a consensus statement. *Stroke* **44**, 2650-2663, doi:10.1161/strokeaha.113.001972 (2013).

26    Higashida, R. T. *et al.* Trial design and reporting standards for intra-arterial cerebral thrombolysis for acute ischemic stroke. *Stroke* **34**, e109-137, doi:10.1161/01.Str.0000082721.62796.09 (2003).

27    Gorgolewski, K. J. *et al.* The brain imaging data structure, a format for organizing and describing outputs of neuroimaging experiments. *Sci Data* **3**, 160044, doi:10.1038/sdata.2016.44 (2016).

28    Wasserthal, J. *et al.* TotalSegmentator: Robust Segmentation of 104 Anatomic Structures in CT Images. *Radiol Artif Intell* **5**, e230024, doi:10.1148/ryai.230024 (2023).

29    de la Rosa, E. et al. AIFNet: Automatic vascular function estimation for perfusion analysis using deep learning. *Med Image Anal* **74**, 102211, doi:10.1016/j.media.2021.102211 (2021).

30    de la Rosa, E. et al. Detecting CTP truncation artifacts in acute stroke imaging from the arterial input and the vascular output functions. *PLoS One* **18**, e0283610, doi:10.1371/journal.pone.0283610 (2023).

31    Klein, S., Staring, M., Murphy, K., Viergever, M. A. & Pluim, J. P. elastix: a toolbox for intensity-based medical image registration. *IEEE Trans Med Imaging* **29**, 196-205, doi:10.1109/tmi.2009.2035616 (2010).

32    Ourselin S, S. R., Pennec X. . Robust registration of multi-modal images: towards real-time clinical applications. . *International Conference on Medical Image Computing and Computer-Assisted Intervention 2002 Sep 25 (pp. 140-147). Berlin, Heidelberg: Springer Berlin Heidelberg.*

33    Isensee, F. *et al.* Automated brain extraction of multisequence MRI using artificial neural networks. *Hum Brain Mapp* **40**, 4952-4964, doi:10.1002/hbm.24750 (2019).

34    Yushkevich, P. A. *et al.* User-guided 3D active contour segmentation of anatomical structures: significantly improved efficiency and reliability. *Neuroimage* **31**, 1116-1128, doi:10.1016/j.neuroimage.2006.01.015 (2006).

35    de la Rosa, E. et al. A Robust Ensemble Algorithm for Ischemic Stroke Lesion Segmentation: Generalizability and Clinical Utility Beyond the ISLES Challenge. *arXiv:2403.19425v2* (2024).

36    Ronneberger, O. F., P.; Brox T. U-Net: Convolutional Networks for Biomedical Image Segmentation. *arXiv:1505.04597* (2015).

37    Kirchhoff, Y. R., M. R.; Roy, S.; Kovacs, B.; Ulrich, C.; Wald, T.; Zenk, M.; Vollmuth, P.; Kleesiek, J.; Isensee, F.; Maier-Hein, K. Skeleton Recall Loss for Connectivity Conserving and Resource Efficient Segmentation of Thin Tubular Structures. *arXiv:2404.03010* (2024).

38    Yang, K. et al. Benchmarking the CoW with the TopCoW Challenge: Topology-Aware Anatomical Segmentation of the Circle of Willis for CTA and MRA. *arXiv:2312.17670* (2024).

39    Silversmith, W. Connected components on multilabel 3D & 2D images. *Zenodo* (2021).